\documentclass{article}

\usepackage{arxiv}

\usepackage[utf8]{inputenc}   
\usepackage[T1]{fontenc}      
\usepackage{amsmath,amssymb,amsfonts} 
\usepackage{cite}             
\usepackage{hyperref}         
\usepackage{url}              
\usepackage{booktabs}         
\usepackage{tabularx}         
\usepackage{array}
\usepackage{multirow}
\usepackage{nicefrac}         
\usepackage{microtype}        
\usepackage{graphicx}
\usepackage{xcolor}
\graphicspath{{./}}

\newcolumntype{Y}{>{\centering\arraybackslash}X}
\newcommand{\modelLogo}[1]{%
  \makebox[1.85em][c]{%
    \includegraphics[width=1.65em,height=1.65em,keepaspectratio]{#1}}%
}
\newcommand{\modelBlock}[2]{%
  \shortstack[c]{\modelLogo{#1}\\[-1pt]\scriptsize\strut #2}%
}
\hypersetup{
  colorlinks = true,
  linkcolor  = {black},
  citecolor  = {black},
  urlcolor   = {blue!60!black}
}

\title{SAGE: A Statistical Acceptance Gate for Self-Evolving Agents}

\author{
\bfseries
Yihao Wang$^{1,\dagger}$\quad
Linhan Xia$^{2,5,\dagger}$\quad
Rui Liu$^{3}$\quad
Zhaofeng Zhang$^{4,6}$\quad
Hongyu Wu$^{2}$\\[3pt]
\bfseries
Yang Yang$^{7}$\quad
Jinglu He$^{7}$\quad
Yu Guo$^{8}$\quad
Kai Lei$^{1,*}$\\[8pt]
\mdseries\small
$^{1}$Peking University \quad
$^{2}$University of Oklahoma \quad
$^{3}$Imperial College London\\[1pt]
\mdseries\small
$^{4}$University of Michigan \quad
$^{5}$Tencent \quad
$^{6}$University of Edinburgh\\[1pt]
\mdseries\small
$^{7}$Xunce Technology \quad
$^{8}$GienTech Technology\\[1pt]
\mdseries\small
$^{\dagger}$These authors contributed equally to this work.\\
$^{*}$Corresponding author: \texttt{leik@pkusz.edu.cn}
}

\begin{document}
\raggedbottom 
\maketitle

\begin{abstract}
Large Language Model (LLM)-based agents increasingly self-evolve by editing a persistent skill document that encodes their workflow, tool-use rules, and decision logic. This loop has two steps, an optimizer that proposes a candidate edit and a gate that accepts or rejects it. Prior work has concentrated on the optimizer, while the gate still follows a naive rule that keeps any edit which improves an aggregate validation score. We show that this rule fails in two ways. First, it admits permanent regressions, since an edit can raise the average while breaking items the skill already solves. Second, it is vulnerable to the Optimizer's Curse, since the best observed score on a finite and noisy validation set is upward biased. To solve the above two limitations, we propose a statistical acceptance gate for self-evolving agents (SAGE). Compared with previous work, SAGE has two contributions. First, SAGE proposes a per-item paired comparison that evaluates the current skill and the edited skill on identical validation items, which exposes regressions that an aggregate score hides and penalizes them asymmetrically. Second, SAGE also employs a one-sided paired test that commits an edit only when its wins are statistically reliable against its losses, and it abstains otherwise. SAGE is a conservative refinement of the standard gate that recovers the baseline exactly at a boundary setting. It commits only a subset of the baseline's edits, filtering out those whose gains are unreliable or purchased by breaking already-solved items. Across five benchmarks and four backbone LLMs under an equal-budget protocol, SAGE lowers the regression rate in 19 of 20 settings and matches the baseline in the remaining one, for example from 36.5\% to 0\% on LiveMath and from 42.8\% to 0\% on OfficeQA with DeepSeek-V4. SAGE also attains the highest final score in all 20 settings, raising LiveMath from 34.15 to 48.78.
\end{abstract}

\section{Introduction}\label{sec:intro}
In the concept of LLM-agent self-evolving, agents increasingly improve themselves by editing a persistent skill document, which is a reusable artifact that encodes their workflow and decision logic \cite{voyager2023,trace2skill2026}. The current state-of-the-art (SOTA), SkillOpt \cite{yang2026skillopt}, treats this document as the trainable external state of a frozen agent and edits it with a disciplined optimizer. They reported strong results across mainstream benchmarks. Such a loop has two steps. First, an optimizer proposes a candidate edit. Second, a gate decides whether to accept it. Almost all methodological attention has gone to the first step, namely how to propose better edits.

Compared with prior work, SkillOpt \cite{yang2026skillopt} addresses a core challenge in document-based self-evolution. The optimization process in the previous works was left unconstrained. Left unconstrained, repeated self-editing degrades the document rather than improving it. SkillOpt's contribution is to bring an optimization discipline to the editing process, so that revisions refine the document rather than destabilize it.

\begin{figure}[t]
  \centering
  \includegraphics[width=\linewidth]{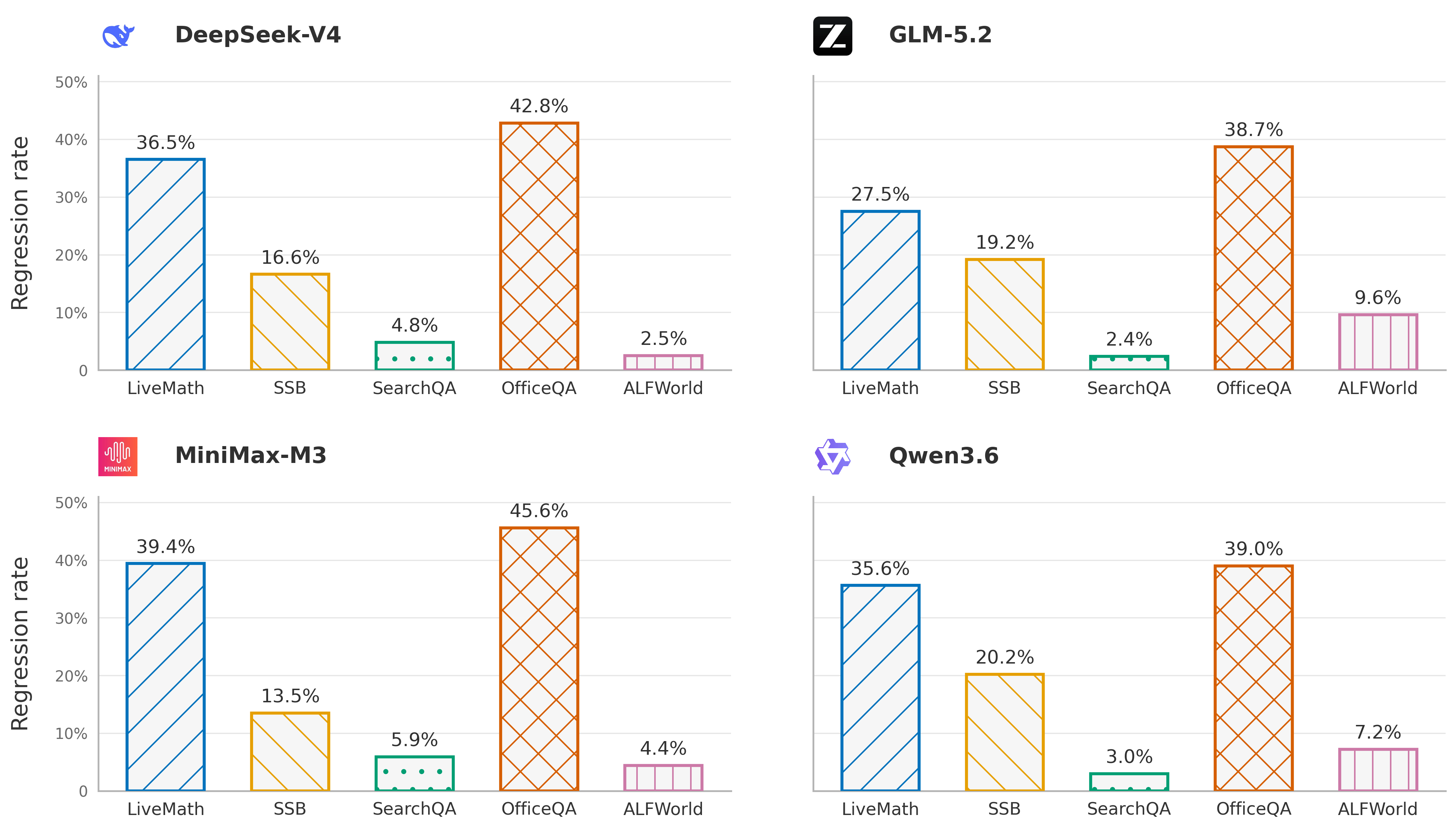}
  \caption{Regression rates of the naive acceptance gate across five
  benchmarks and four backbone LLMs. Each panel corresponds to one
  backbone model. Higher values indicate more frequent regressions.}
  \label{fig:naive-acr}
\end{figure}

However, SkillOpt \cite{yang2026skillopt} focuses on the optimization side rather than the acceptance side. Across all these approaches, even including the current SOTA, the gate follows a naive rule. They keep an edit once it improves the aggregate validation score. This method is useful and practical for optimizing overall validation performance, but there is room for improvement in generalization.

In terms of diving into the details, the aggregate-and-point-estimate rule has two independent weaknesses. The first weakness is that the evidence the gate is based on is insufficient. It only sees the aggregate score among the validation set. An edit can raise the aggregate score while breaking cases that the current skill already solves. Since an accepted edit is irreversible, such a harmful regression is permanently written into the skill and is difficult to modify. These regressions are not a rare problem, as across the five benchmarks we test, the regression rates range from \(2.5\%\) in ALFWorld \cite{shridhar2021alfworld} and \(4.8\%\) in SearchQA \cite{dunn2017searchqa} to \(42.8\%\) in OfficeQA \cite{opsahl2026officeqa}. The entire result is shown in Fig.~\ref{fig:naive-acr}. By trading away mastered behavior for average-case gains, the gate corrupts its own optimization trajectory and settles into a local optimum that is hard to leave.

The second weakness is that previous works do not pay attention to how much the gate trusts that evidence. Validation is finite and noisy. So choosing the edit with the highest observed score is upward biased, a phenomenon known as the Optimizer's Curse \cite{smith2006optimizer}. The gate commits edits that were lucky rather than better. Such lucky edits overfit the validation set and do not transfer, so apparent validation gains fail to reach the test distribution.

To address the two aforementioned weaknesses of the current SOTA, we propose SAGE. It is a gate that recasts acceptance as a statistical decision under noisy validation and conservatively refines the standard gate. SAGE pairs one component to each weakness. A per-item paired evaluation compares the current skill and the edited skill on the same validation items. It causes a harmful local loss to no longer hide inside an average. The second component is a hypothesis-test denoising rule that accepts an edit only when its wins are statistically reliable against its losses. Together the two components make the gate have a wider global view and trust more carefully.

Our contributions are fourfold:
\begin{enumerate}
  \item \textbf{Diagnosis.} We identify the acceptance step of skill
  self-evolution as an overlooked bottleneck. We trace the standard gate's
  failure to two causes. First, it admits permanent regressions. Second, it
  falls to the Optimizer's Curse under a small, noisy validation set.

  \item \textbf{Method.} We propose SAGE, a regression-aware and noise-aware acceptance gate that can abstain. SAGE is a conservative refinement of the standard gate and recovers it exactly at a boundary parameter setting.

  \item \textbf{Paired evidence.} We judge confidence from paired, per-item
  evidence. This avoids a key bias. Even a corrected aggregate score still
  suffers from it.

  \item \textbf{Results.} We test under an equal-budget protocol across
  five benchmarks and four backbone LLMs. The standard gate admits these
  regressions across all tested models. SAGE admits far fewer and attains
  the highest final score in all 20 settings.
\end{enumerate}

\section{Roadmap of Self-Evolving Agents}\label{sec:related}
The earliest form of LLM self-improvement refines a single output or a reasoning trace. Self-Refine \cite{selfrefine2023} iterates a loop within one episode, which includes generation, critique, and revision to optimize the behavior. Reflexion \cite{shinn2023reflexion} turns environment feedback into verbal lessons held in a short-term buffer, and STaR \cite{star2022} bootstraps a model on its own successful rationales. The gains of the first two fade with the episode or the buffer, while only STaR distills them into the model's weights. However, none of these methods maintains a persistent, separately optimized artifact, so their improvements do not accumulate in a transferable form. Besides, the knowledge that does persist is embedded in parameters, which prevents humans from supervising the optimization process to keep it safe.

To address the above-mentioned problem, the community has begun to explore treating natural-language artifacts as objects of optimization. At the very beginning stage, scholars were trying to optimize the prompt. OPRO \cite{opro2024} uses the LLM itself as an optimizer over an instruction. EvoPrompt~\cite{evoprompt2024} and PromptBreeder~\cite{promptbreeder2023} evolve a population of prompts. TextGrad~\cite{textgrad2025}, GEPA~\cite{gepa2025}, and DSPy~\cite{dspy2024} backpropagate textual feedback or compile and tune multi-stage pipelines. With the rapid advance of AI agents, these ideas have been extended from prompt optimization to the optimization of agent skills. Unlike a prompt, a skill is a reusable procedural artifact that encodes workflow, tool-use rules, and decision logic. Voyager \cite{voyager2023} grows a skill library during exploration. ADAS~\cite{adas2024} searches over agent designs expressed in code, and Trace2Skill~\cite{trace2skill2026} consolidates trajectory-local lessons into a skill directory. However, these methods edit the agent artifact without the discipline that makes the optimization stable. Edits are unbounded and updates react directly to raw rollout feedback, which leaves the process noisy and hard to reproduce.

In this line, the current state of the art is SkillOpt~\cite{yang2026skillopt}, which brings this missing discipline to skill editing. It treats the skill as the trainable external state of a frozen agent and updates it like a learned parameter. Each edit is bounded by a textual learning rate, so a single step cannot overwrite too much. Besides, SkillOpt has a rejected-edit buffer that remembers proposals that have already failed, and a slow, epoch-wise update stabilizes the optimization trajectory. With this controlled generator, SkillOpt reports state-of-the-art results across six mainstream benchmarks. However, its acceptance mechanism is naive, an edit is kept once it improves the aggregate validation score. In other words, the validator can be misled by the validation set itself. First, an aggregate score can hide harmful modification, an edit may improve the average performance while degrading important cases or subskills. Second, once an edit is accepted, the gate cannot tell whether the observed gain reflects a genuine improvement or merely a lucky fluctuation on the validation set. Repeating this process can therefore push the optimization toward a locally favorable but not global optimum. This eventually causes the skill to overfit the validation set rather than improve its general behavior. This phenomenon is observed in our experiments and introduced in the Introduction.

This weakness is not specific to SkillOpt. In the methods above, an edit is usually accepted as long as it improves the aggregate validation score. Yet validation is finite and noisy: a higher score may reflect a lucky evaluation rather than a real improvement. Repeating this rule can therefore select edits that look good on validation but do not improve the underlying skill. To our knowledge, the acceptance step in skill self-evolution has not been explicitly studied under evaluation noise, and this is the focus of our work.

\section{Methodology}\label{sec:method}
We formalize the acceptance gate of skill self-evolution. At each step an optimizer proposes a candidate edit and the gate decides whether it replaces the incumbent. We cast this decision as a statistical test under noisy validation, built from two coupled components. First is a per-item paired evaluation, and second is a hypothesis test that denoises the decision.

\subsection{Overview}\label{subsec-method:overview}
Following the setting of SkillOpt \cite{yang2026skillopt}, at step \(t\) the validation gate receives the current skill \(S^{(t)}\) and candidate edits \(E=[e_1,\dots,e_b]\), where \(b\) is the size of the mini-batch denoting the number of candidate edits. The gate scores each candidate on a held-out validation set \(D\) via a frozen
target model and a binary verifier \(v(S,x)\in\{0,1\}\), which indicates whether skill \(S\) solves question \(x\). For each edit \(e_n\), we define the corresponding candidate skill as
\begin{equation}
    \widetilde S_n^{(t+1)}=S^{(t)}+e_n,
    \label{eq:candidate}
\end{equation}
where the tilde distinguishes a candidate skill from the committed incumbent \(S^{(t+1)}\). The baseline gate decides to keep or discard each edit by its aggregate score,
\begin{equation}
    J_D\!\left(\widetilde S_n^{(t+1)}\right)=\tfrac{1}{|D|}\sum_{x\in D} v\!\left(\widetilde S_n^{(t+1)},x\right),
    \label{eq:baseline}
\end{equation}
where \(J_D(S^{(\cdot)})\) is the score of skill \(S^{(\cdot)}\) over question set \(D\). It keeps an edit exactly when its score strictly improves over the current skill,
\begin{equation}
    g_0(e_n, S^{(t)})= I\!\left[\,J_D\!\left(\widetilde S_n^{(t+1)}\right) > J_D\!\left(S^{(t)}\right)\,\right],
  \label{eq:skillopt}
\end{equation}
where \(g_0\) is the accept gate of the SkillOpt baseline and \(I\) is a binary function to indicate acceptance of the edit.

\begin{figure}[t]
  \centering
  \includegraphics[width=\linewidth]{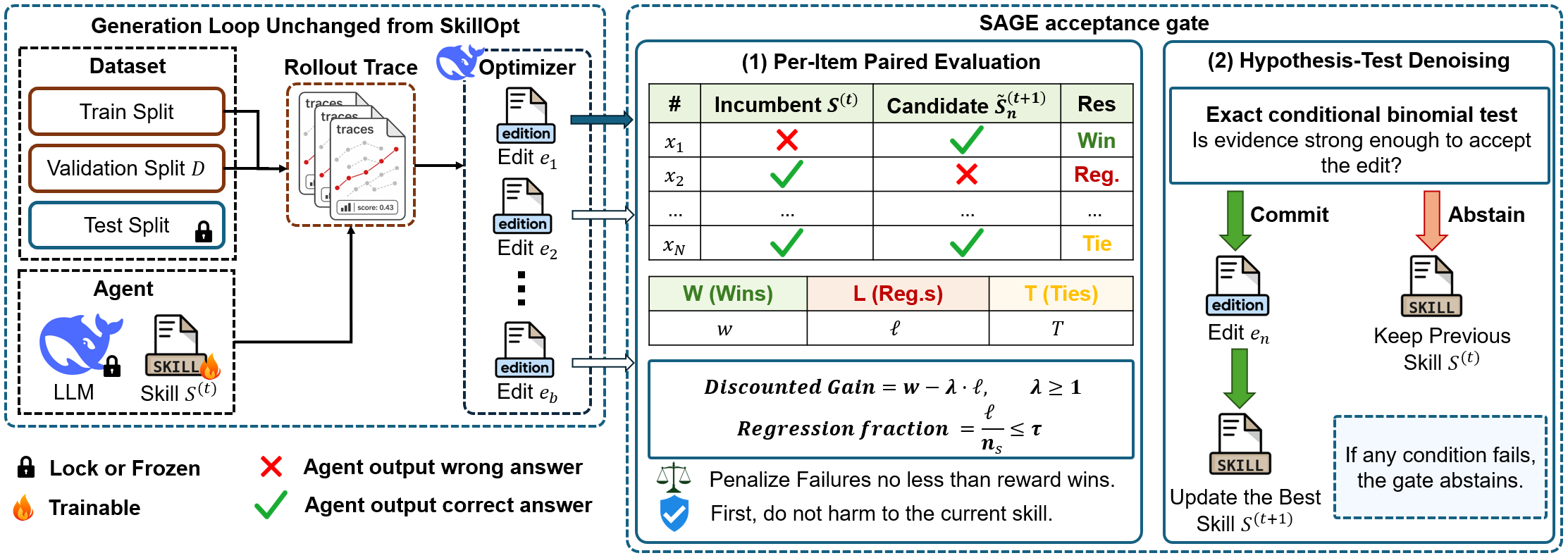}
  \caption{The pipeline of SAGE. ``Reg.'' abbreviates ``Regression''. Each edit $e_n$ yields a candidate skill $\widetilde S_n^{(t+1)} = S^{(t)}+e_n$. The gate commits it as the new incumbent only when $\Delta_\lambda>0$, $p_\lambda(w,\ell)<\alpha$, and $\rho\le\tau$, and otherwise retains $S^{(t)}$. The diagram uses DeepSeek as an illustrative backbone. Our experiments cover four backbone LLMs.}
  \label{fig:gate}
\end{figure}

Under finite and noisy validation, \(g_0\) may keep edits that are lucky rather than better, which is called the Optimizer's Curse \cite{smith2006optimizer} and has been discussed in the Introduction. In the setting of agent self-evolving, it comes from two reasons introduced in the Introduction, and we address them with two newly proposed components. The first reason is that the baseline reads only the aggregate score. It is blind to which items an edit fixes versus breaks. SAGE replaces this with a per-item paired comparison that also penalizes regressions more than repairs. The second reason is that the baseline trusts the point value of the score, with no guard against the sampling noise of the validation set. SAGE keeps an edit only when the paired evidence is unlikely to arise from that noise alone, and otherwise abstains. The two components share one per-item ledger, and the test in the second runs on the paired data produced by the first. The design is faithful to its baseline: disabling both recovers \(g_0\) exactly, as shown formally in the Methodology and verified in the ablation study. The process is demonstrated in Fig.~\ref{fig:gate}.

\subsection{Per-Item Paired Evaluation}\label{sec:paired}
Rather than reducing an edit to one aggregate number across the various tasks \(x\in D\), we compare each edit \(e_n\) with the current skill \(S^{(t)}\) on each validation item \(x\in D\). We run the incumbent \(S^{(t)}\) and the candidate skill \(\widetilde S_n^{(t+1)}\) defined in Eq.~\eqref{eq:candidate} on the same items \(x\in D\). We classify each item as a win (the edit solves what the current skill misses), a regression (the current skill solves it, the edit breaks it), or a tie. Tied items that the two agree on carry no comparative signal. For a fixed candidate \(e_n\), each validation item belongs to one of four paired outcomes: both correct, both incorrect, a win, or a regression. Only the two discordant outcomes carry comparative information. Let
\begin{equation}
    w=\lvert\{x\in D: v(\widetilde S_n^{(t+1)},x)=1,\,v(S^{(t)},x)=0\}\rvert,
\end{equation}
count wins and
\begin{equation}
    \ell=\lvert\{x\in D: v(\widetilde S_n^{(t+1)},x)=0,\,v(S^{(t)},x)=1\}\rvert
\end{equation}
count regressions. Evaluating on identical items makes each item a head-to-head contest. The aggregate score is a lossy projection of this ledger: an edit with \(w=\ell+1\) improves the mean score and is accepted by \(g_0\), but it may have quietly broken \(\ell\) items the current skill already solves. It is a trade that \(g_0\) cannot see but the paired counts expose.

Because a committed skill is deployed irreversibly, breaking a solved item is more harmful than missing a new one. So we weight the two asymmetrically. Let
\begin{equation}
  n_S=\lvert\{x\in D: v(S^{(t)},x)=1\}\rvert
  \label{eq:ns}
\end{equation}
be the number of validation items solved by the incumbent. We summarize the ledger by a discounted net gain and an empirical regression fraction,
\begin{equation}
  \Delta_\lambda \;=\; w - \lambda\,\ell,
  \qquad
  \rho \;=\;
  \begin{cases}
    \ell/n_S, & n_S>0,\\
    0, & n_S=0,
  \end{cases}
  \label{eq:delta}
\end{equation}
where \(\lambda\ge 1\) makes a regression cost at least as much as a repair rewards. It follows the first-do-no-harm principle of asymmetric counterfactual utilities~\cite{benmichael2024,predictionupdate2021}, and \(\rho\) caps the fraction of solved items an edit may break. The case \(n_S=0\) forces \(\ell=0\), so the convention \(\rho=0\) is harmless. In our setting, both wins and regressions are directly observed through paired evaluation of the incumbent and candidate on the same validation items. The hyperparameters \(\lambda\), \(\tau\), and \(\alpha\) are tuned on held-out data only. The test set is unseen during the whole optimization process.

\subsection{Hypothesis-Test Denoising}\label{sec:denoise}
The first component is responsible for yielding the evidence, while the second component decides how much of it to trust. This distinction is necessary because \(\Delta_\lambda\) is computed from a finite validation set and is therefore noisy. If we evaluate the same pair of skills on a different sample of items, the measured improvement may change.

For this reason, SAGE does not accept a new edit simply because \(\Delta_\lambda > 0\). A positive estimate may reflect a real improvement, but it may also be caused by sampling noise. What matters is whether the observed gain is large and stable enough relative to its uncertainty. The gate therefore requires the improvement to be positive with statistical confidence.

This confidence requirement reduces the chance of committing lucky but unreliable edits. For example, two candidate edits may obtain the same value of $\Delta_\lambda$. One may achieve it through a few wins with no regressions, while the other may accumulate many wins that are largely offset by regressions. The latter is more likely to be a false winner: its apparent improvement is driven by luck rather than a robust change in skill quality. Accepting such candidates is a form of the Optimizer's Curse, because repeated search tends to favor edits whose estimates are inflated by noise.

The confidence must be measured on the paired ledger, not by denoising the aggregate score. Shrinking each edit's aggregate score toward the mean is the natural fix and is directionally correct \cite{smith2006optimizer}. However, on a shared validation set the edits' scores are correlated through common item difficulty, and shrinking correlated aggregates fails twice. First, the shrinkage factor is mis-estimated under that correlation. Second, reducing the error of each score does not imply a more accurate ranking \cite{winnerscurse2023}. The paired view controls for the confound directly because the current skill and the edit face the same item, so shared item difficulty is held fixed within each comparison, although stochastic execution noise may remain. We therefore test the discordant pairs against the break-even level implied by the asymmetric cost. For a fixed incumbent--candidate pair, let \(W\) denote the event that the candidate solves a task instance drawn from the target task distribution while the incumbent does not, and let \(L\) denote the reverse event. If model execution is stochastic, its randomness is part of this probability space. The population \(\lambda\)-weighted gain is
\begin{equation}
  G_\lambda \;=\; \Pr(W)-\lambda\,\Pr(L).
  \label{eq:popgain}
\end{equation}
Writing \(q=\Pr(W\mid W\cup L)\) for the probability that a discordant outcome is a win, we have \(G_\lambda=\Pr(W\cup L)\bigl((1+\lambda)q-\lambda\bigr)\) whenever \(\Pr(W\cup L)>0\), so a positive population gain is equivalent to
\begin{equation}
  q>q_\lambda,
  \qquad
  q_\lambda \;=\; \frac{\lambda}{1+\lambda}.
  \label{eq:qlambda}
\end{equation}
The observed ledger obeys the same geometry: with \(m=w+\ell\) and \(\widehat q=w/m\) for \(m>0\), the identity \(\Delta_\lambda=m\bigl((1+\lambda)\widehat q-\lambda\bigr)\) holds, so \(\Delta_\lambda>0\) exactly when \(\widehat q>q_\lambda\). We test \(H_0\!:q\le q_\lambda\) against \(H_1\!:q>q_\lambda\). Conditional on the observed number \(m\) of discordant pairs, the one-sided exact significance is
\begin{equation}
  p_\lambda(w,\ell)
   \;=\;
  \Pr\!\left[\,\mathrm{Bin}\!\left(m,\,q_\lambda\right) \ge w\,\right],
  \label{eq:ptest}
\end{equation}
which is valid for the composite null because the binomial upper tail is monotone in \(q\). When \(m=0\), the formula gives \(p_\lambda=1\), so the gate abstains without comparative evidence. This is a one-sided exact conditional binomial test on the discordant pairs. For \(\lambda=1\), it reduces to the one-sided exact McNemar test, whose null treats each discordant pair as a fair coin flip. As \(\lambda\) grows, the threshold \(q_\lambda\) rises, so a candidate must show stronger win dominance to offset each regression. In all cases, what governs the decision is the composition of wins and losses, not the net count: with \(\lambda=1\), a record of five wins to three losses does not reach significance, whereas eight wins to zero losses does.

The accept rule combines the positive-gain requirement, this test, and the regression cap. We commit the edit, setting \(S^{(t+1)}=\widetilde S_n^{(t+1)}\), only when the empirical discounted gain is positive, the paired test rejects at level \(\alpha\), and the regression rate stays within \(\tau\). Otherwise the gate abstains and retains \(S^{(t+1)}=S^{(t)}\):
\begin{equation}
    \mathcal{A}(e_n, S^{(t)}) =
    I\!\left[\,\Delta_\lambda>0,\ p_\lambda(w,\ell) < \alpha,\ \rho \le \tau\,\right].
\label{eq:gate}
\end{equation}

The gate family in Eq.~\eqref{eq:gate} contains the SkillOpt baseline as a boundary member. At \((\lambda,\alpha,\tau)=(1,1,1)\), the significance and cap conditions hold whenever \(\Delta_1>0\), so the gate accepts exactly when \(w-\ell>0\), which is the baseline rule \(g_0\) because \(J_D(\widetilde S_n^{(t+1)})-J_D(S^{(t)})=(w-\ell)/|D|\). Under a conservative setting with \(\lambda\ge 1\), \(\alpha\le 1\), and \(\tau\le 1\), the gate is instead a refinement of \(g_0\): any accepted edit has \(w>\lambda\ell\ge\ell\), so it would also be accepted by \(g_0\).

The level \(\alpha\) controls how conservative the gate is. It can be interpreted either as a one-sided significance threshold or as requiring a one-sided lower confidence bound for $q$ to exceed $q_\lambda$. In practice, \(\alpha\) should match the task noise. For instance, verifiable tasks can use a looser gate, while noisy tasks should use a stricter one, where abstention is expected.

\section{Experimental Settings and Analysis}\label{sec:experiment}

\subsection{Setup}
To evaluate the proposed acceptance gate, and following the setting of SkillOpt \cite{yang2026skillopt}, we evaluate on five benchmarks. These benchmarks include LiveMath \cite{he2026livemathematicianbench}, SpreadsheetBench (SSB) \cite{ma2024spreadsheetbench}, SearchQA \cite{dunn2017searchqa}, OfficeQA Pro (abbreviated as OfficeQA) \cite{opsahl2026officeqa}, and ALFWorld \cite{shridhar2021alfworld}. We did not test on DocVQA \cite{mathew2021docvqa} because not all four backbones support visual input, and we keep the evaluation protocol identical across models. Extending SAGE to multimodal LLMs is left for future work. To isolate the acceptance decision, we hold generation fixed to the setting of SkillOpt. This means every edit is proposed by the same SkillOpt optimizer and the only difference is the gate that accepts them. For better fairness, we use a matched budget for both the baseline and the SAGE-integrated method, which means an equal number of proposed edits and equal token cost. To enhance the reproducibility of SAGE, we take an open-source model as the base model. We use DeepSeek-V4-flash, GLM-5.2, MiniMax-M3, and Qwen3.6 as the base models for both vanilla SkillOpt and the proposed SAGE-integrated method, which are one of the most advanced open-source models.

\subsection{SAGE Reduces Regressions}
We first ask whether SAGE commits fewer regressions than the naive gate. The regression rate is the fraction of the items the current skill already solves that an accepted edit breaks. For this metric, lower is better. Table~\ref{tab:acr} reports this rate for both gates. SAGE lowers the regression rate in 19 of the 20 backbone--benchmark settings and matches the baseline in the remaining one (GLM-5.2 on SearchQA). With DeepSeek-V4, the rate drops from \(36.5\%\) to \(0\%\) on LiveMath and from \(16.6\%\) to \(0\%\) on SpreadsheetBench. On SearchQA the naive gate already regresses little (\(4.8\%\) for DeepSeek-V4) and the remaining gap is small. On OfficeQA, the regression rate drops from \(42.8\%\) to \(0\%\). On ALFWorld the naive gate regresses comparatively little across backbones, and SAGE further reduces or matches the rate in every case. Similar to SearchQA, there is less to prevent, and SAGE remains safe on this axis.

\begin{table*}[t]
\setlength{\abovecaptionskip}{0pt}
\setlength{\belowcaptionskip}{7pt}
\centering
\caption{Regression rate across five benchmarks and four backbone LLMs.
Lower is better. The lower regression rate within each backbone and
benchmark is shown in \textbf{boldface}.}
\label{tab:acr}
\small
\setlength{\tabcolsep}{4pt}
\renewcommand{\arraystretch}{1.15}

\begin{tabular*}{\textwidth}{@{\extracolsep{\fill}}>{\centering\arraybackslash}m{1.9cm}lccccc@{}}
\toprule
Backbone & Gate
& LiveMath & SSB & SearchQA & OfficeQA & ALFWorld \\
\midrule

\multirow{2}{*}{\modelBlock{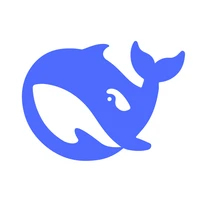}{DeepSeek-V4}}
& Naive \(g_0\)
& 36.5\% & 16.6\% & 4.8\% & 42.8\% & 2.5\% \\
& SAGE
& \textbf{0.0\%} & \textbf{0.0\%} & \textbf{3.7\%}
& \textbf{0.0\%} & \textbf{1.9\%} \\
\cmidrule(lr){1-7}

\multirow{2}{*}{\modelBlock{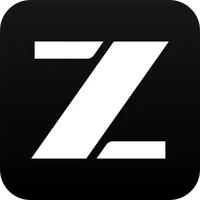}{GLM-5.2}}
& Naive \(g_0\)
& 27.5\% & 19.2\% & 2.4\% & 38.7\% & 9.6\% \\
& SAGE
& \textbf{0.0\%} & \textbf{0.0\%} & 2.4\% & \textbf{0.0\%} & \textbf{5.1\%} \\
\cmidrule(lr){1-7}

\multirow{2}{*}{\modelBlock{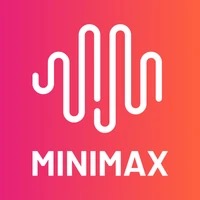}{MiniMax-M3}}
& Naive \(g_0\)
& 39.4\% & 13.5\% & 5.9\% & 45.6\% & 4.4\% \\
& SAGE
& \textbf{0.0\%} & \textbf{0.0\%} & \textbf{1.3\%} & \textbf{0.0\%} & \textbf{0.0\%} \\
\cmidrule(lr){1-7}

\multirow{2}{*}{\modelBlock{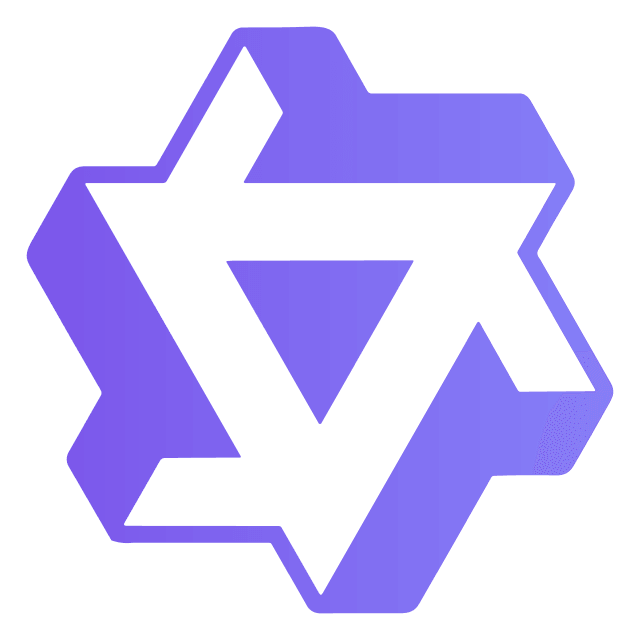}{Qwen3.6}}
& Naive \(g_0\)
& 35.6\% & 20.2\% & 3.0\% & 39.0\% & 7.2\% \\
& SAGE
& \textbf{0.0\%} & \textbf{0.0\%} & \textbf{2.7\%} & \textbf{0.0\%} & \textbf{4.4\%} \\
\bottomrule
\end{tabular*}
\end{table*}

\subsection{SAGE Achieves the Best Final Performance}
We next compare the final committed skills across four backbone LLMs and isolate the contribution of each component. Table~\ref{tab:main} reports vanilla SkillOpt, the paired-comparison variant (C1), which uses the per-item paired criterion without the statistical test, and the full SAGE method across five benchmarks. Within every backbone and benchmark, SAGE achieves the highest final score. Across all 20 backbone--benchmark pairs, SAGE improves over vanilla SkillOpt by \(8.73\) points on average and its reported final score is the highest in every setting.

The trend is consistent across models. For DeepSeek-V4, SAGE raises the final score from \(34.15\) to \(48.78\) on LiveMath and from \(32.93\) to \(45.12\) on OfficeQA, while improving all remaining benchmarks as well. GLM-5.2 gains \(13.60\) points on LiveMath and \(11.73\) points on OfficeQA. MiniMax-M3 exhibits the largest improvements, with gains of \(16.61\) points on LiveMath and \(16.93\) points on OfficeQA. Qwen3.6 also improves on every benchmark, including a \(16.80\)-point gain on LiveMath and a \(10.32\)-point gain on OfficeQA. These results show that the benefit of SAGE is not specific to one backbone or one task.

Table~\ref{tab:main} also shows that the two components are complementary. C1 improves over vanilla SkillOpt in every setting, contributing an average gain of \(6.20\) points. The full SAGE gate further improves over C1 in all 20 comparisons, adding \(2.53\) points on average. Thus, the paired comparison provides a substantial regression-aware improvement, while the statistical acceptance rule consistently delivers additional gains beyond C1.

\begin{table*}[t]
\setlength{\abovecaptionskip}{0pt}
\setlength{\belowcaptionskip}{7pt}
\centering
\caption{Final scores across five benchmarks and four backbone LLMs.
The best score within each backbone and benchmark is shown in \textbf{boldface}.}
\label{tab:main}
\small
\setlength{\tabcolsep}{4pt}
\renewcommand{\arraystretch}{1.15}

\begin{tabular*}{\textwidth}{@{\extracolsep{\fill}}>{\centering\arraybackslash}m{1.9cm}lccccc@{}}
\toprule
Backbone & Method
& LiveMath & SSB & SearchQA & OfficeQA & ALFWorld \\
\midrule

\multirow{3}{*}{\modelBlock{deepseek.png}{DeepSeek-V4}}
& SkillOpt
& 34.15 & 47.56 & 81.71 & 32.93 & 66.58 \\
& \quad + paired (C1)
& 46.58 & 50.97 & 82.44 & 40.48 & 69.92 \\
& \quad + SAGE
& \textbf{48.78} & \textbf{51.22} & \textbf{84.15}
& \textbf{45.12} & \textbf{72.92} \\
\cmidrule(lr){1-7}

\multirow{3}{*}{\modelBlock{glm.png}{GLM-5.2}}
& SkillOpt
& 42.12 & 59.74 & 77.20 & 44.46 & 69.89 \\
& \quad + paired (C1)
& 51.50 & 60.33 & 82.13 & 52.75 & 74.10 \\
& \quad + SAGE
& \textbf{55.72} & \textbf{61.42} & \textbf{85.57}
& \textbf{56.19} & \textbf{77.76} \\
\cmidrule(lr){1-7}

\multirow{3}{*}{\modelBlock{minimax.jpg}{MiniMax-M3}}
& SkillOpt
& 30.97 & 51.73 & 80.20 & 43.19 & 62.88 \\
& \quad + paired (C1)
& 45.47 & 52.10 & 84.68 & 55.79 & 72.41 \\
& \quad + SAGE
& \textbf{47.58} & \textbf{54.14} & \textbf{87.07}
& \textbf{60.12} & \textbf{74.38} \\
\cmidrule(lr){1-7}

\multirow{3}{*}{\modelBlock{qwen.png}{Qwen3.6}}
& SkillOpt
& 40.30 & 47.64 & 69.57 & 37.07 & 71.30 \\
& \quad + paired (C1)
& 53.34 & 49.02 & 74.79 & 44.28 & 72.09 \\
& \quad + SAGE
& \textbf{57.10} & \textbf{50.24} & \textbf{76.08}
& \textbf{47.39} & \textbf{72.91} \\
\bottomrule
\end{tabular*}
\end{table*}

\section{Discussion}\label{sec:discussion}

\subsection{The Benefit of SAGE Scales with Regression Risk}\label{subsec:scope}
The gain from SAGE is not uniform across tasks, and its magnitude tracks how readily the baseline gate regresses on each benchmark. On the two benchmarks where the naive gate regresses most, namely LiveMath at \(36.5\%\) and OfficeQA at \(42.8\%\), SAGE removes regressions entirely and also delivers the two largest score gains, raising LiveMath from \(34.15\) to \(48.78\) and OfficeQA from \(32.93\) to \(45.12\) in final score. On the benchmarks where the naive gate already regresses little, namely ALFWorld at \(2.5\%\) and SearchQA at \(4.8\%\), the benefit is small. The benefit of SAGE therefore grows with the regression risk that a task presents to the gate.

The reason for this pattern is that the gate exists to intercept harmful edits, so a task that produces few harmful edits leaves little for any gate to intercept. This separates the benchmarks into two regimes. Verifiable and low-noise tasks give the optimizer little room to regress, and there SAGE reduces to a near pass-through that preserves the baseline behavior. Noisy tasks that invite regressions are where the statistical gate does its real work. A gate should act in exactly this way, intervening when an edit threatens mastered behavior and staying inert otherwise.

\subsection{Limitations}\label{subsec:limitations}
We state the assumptions and limits of SAGE honestly. The method assumes a binary verifier with \(v\in\{0,1\}\), and a continuous or graded reward would require a different paired statistic such as a signed-rank test that the current gate does not cover. The sign test on the paired ledger uses only win and loss counts and discards magnitude, so it can be insensitive to a small but consistent real improvement. The statistical guarantee is also per-comparison in scope: the exact test controls the false-commit probability for a fixed incumbent--candidate pair, but the validation set is reused across optimization steps, so trajectory-level error rates are not controlled.

The evaluation has practical limits as well. The hyperparameters \(\lambda\), \(\tau\), and \(\alpha\) are still tuned on held-out data. As a result, the gate is not yet free of manual configuration. The coverage is also partial. Not all four backbones support visual input, so we restrict the evaluation to text-only benchmarks and omit DocVQA to keep the protocol identical across models, which leaves multimodal generalization open. Each of these limits marks a boundary that the next subsection turns into concrete directions.

\subsection{Future Work}\label{subsec:future}
The limitations above point to several extensions of the method. A first direction is a paired statistic for continuous or graded verifiers, for example a signed-rank test \cite{wilcoxon1945individual}, so that the gate applies beyond the binary reward setting. A second direction is an adaptive threshold that tightens or loosens \(\alpha\) from an online estimate of task noise, drawing on variance-aware or time-uniform sequential confidence methods \cite{maurer2009empirical,howard2021time}. This would remove manual tuning and realize the principle that the level should match task noise. A third direction is a co-design of the optimizer and the gate rather than the decoupled setting of this work, where generation is fixed and only the gate varies. Recent work on optimizing language-model pipelines suggests that prompt or module generation and evaluation can be treated as parts of the same optimization loop~\cite{dspy2024}, which motivates jointly optimizing the proposal mechanism and the acceptance rule.

The application scope can also widen. One step is multimodal evaluation that adds DocVQA \cite{mathew2021docvqa} and a multimodal LLM such as LLaVA \cite{liu2023visual}, which would test whether the gate is modality agnostic. A larger step is the move from a single agent with one skill document to multi-agent systems in which several agents maintain and evolve their own documents. Existing LLM-agent systems already show that communication protocols and structured workflows can enable multi-agent collaboration \cite{li2023camel,hong2024metagpt}. The acceptance problem changes in this setting, because the gain or regression of one edit is no longer decidable from local evidence alone, and a local win or loss can be amplified or cancelled at the team level once agents interact. Per-item paired comparison and statistical acceptance would then need to extend from a single skill to a joint acceptance rule that accounts for inter-agent dependence and credit assignment, a long-standing difficulty in cooperative multi-agent learning \cite{foerster2018counterfactual}. Specific questions follow, including how a harmful edit propagates along a collaboration chain to corrupt other agents and how several gates coordinate their decisions under a shared budget. Multi-agent acceptance is a natural and harder extension of the view taken here, in which the gate protects the stability of the whole collaborating system rather than a single skill.

\section{Conclusion}\label{sec:sonclusion}
This work identified the acceptance step of skill self-evolution as an overlooked bottleneck. The standard gate keeps any edit that improves an aggregate validation score, and we traced two failures of this rule. It admits permanent regressions that an average can hide, and it falls to the Optimizer's Curse on a finite and noisy validation set. We proposed SAGE, a statistical acceptance gate that pairs one safeguard to each failure. A per-item paired comparison exposes regressions that the aggregate score conceals and penalizes them asymmetrically, and a one-sided paired test commits an edit only when its wins are reliable against its losses and abstains otherwise. SAGE is a conservative refinement of the standard gate and recovers it exactly at a boundary setting of its parameters.

Experiments on five benchmarks and four backbone LLMs under an equal-budget protocol support these claims. SAGE lowers the regression rate in 19 of 20 settings and matches the baseline in the remaining one, for example from \(36.5\%\) to \(0\%\) on LiveMath and from \(42.8\%\) to \(0\%\) on OfficeQA with DeepSeek-V4, and its reported final score is the highest in all 20 settings, raising LiveMath from 34.15 to 48.78. The benefit is largest on tasks where the baseline gate regresses most and shrinks to a safe no-op where it does not. These results show that acceptance deserves the same methodological attention that prior work has given to edit proposal, and they suggest that statistical acceptance extends naturally to graded verifiers and to multi-agent settings where edits interact across agents.

\bibliographystyle{unsrt}
\bibliography{references}

\end{document}